# A One Stop 3D Target Reconstruction and multilevel Segmentation Method

Jiexiong Xu[1], Weikun Zhao[1], Zhiyan Tang[1] and Xiangchao Gan[1,2*]

[1]*State Key Laboratory of Crop Genetics & Germplasm Enhancement and Utilization, Jiangsu Nanjing Rice Germplasm Resources National Field Observation and Research Station，Jiangsu Engineering Research Center for Plant Genome Editing, Bioinformatics Center, Academy for Advanced Interdisciplinary Studies, College of Engineering, Nanjing Agricultural University, Nanjing 210095, China*

[2]*Zhongshan Biological Breeding Laboratory，Nanjing 210095, China*

[*] Author for correspondence: e-mail address (gan@njau.edu.cn)

**Abstract**

3D object reconstruction and multilevel segmentation are fundamental to computer vision research. Existing algorithms usually perform 3D scene reconstruction and target objects segmentation independently, and the performance is not fully guaranteed due to the challenge of the 3D segmentation. Here we propose an open-source one stop 3D target reconstruction and multilevel segmentation framework (OSTRA), which performs segmentation on 2D images, tracks multiple instances with segmentation labels in the image sequence, and then reconstructs labelled 3D objects or multiple parts with Multi-View Stereo (MVS) or RGBD-based 3D reconstruction methods. We extend object tracking and 3D reconstruction algorithms to support continuous segmentation labels to leverage the advances in the 2D image segmentation, especially the Segment-Anything Model (SAM) which uses the pretrained neural network without additional training for new scenes, for 3D object segmentation. OSTRA supports most popular 3D object models including point cloud, mesh and voxel, and achieves high performance for semantic segmentation, instance segmentation and part segmentation on several 3D datasets. It even surpasses the manual segmentation in scenes with complex structures and occlusions. Our method opens up a new avenue for reconstructing 3D targets embedded with rich multi-scale segmentation information in complex scenes. OSTRA is available from https://github.com/ganlab/OSTRA.

**INTRODUCTION**

Reconstruction and segmentation of 3D objects play a fundamental role in computer vision and have wide applications in object detection and recognition[1]–[4], scene analysis and understanding[5], [6], medical image analysis[7], [8], and auto piloting[10]. According to the level of granularity of segmentation, 3D data segmentation methods can be classified into three categories: semantic segmentation (scene level), instance segmentation (object level), and part segmentation (part level)[9]. 3D objects are usually encoded in different data models, such as point cloud, mesh and voxel, depending on the data acquisition method. Each of model requires the specific 3D segmentation method.

There are heated research interests in 3D data segmentation, and most algorithms focus on the deep learning model. Unfortunately, there are many challenges for these algorithms. The first is that these methods typically require a large amount of annotated data for training, especially for complex segmentation tasks involving diverse scenes. Capturing and annotating large amounts of 3D data is a time-consuming and costly process. At present, there are only a very limited number of datasets are available and the scarcity of annotated datasets impedes the applicability and scalability of these methods. The second challenge stems from dense feature representations which the regular deep learning based methods rely on. Since 3D data which is usually sparse and incomplete, deep learning based algorithm could struggle with 3D segmentation especially during the training process. In addition, due to the "curse of dimensionality", the complexity of 3D object segmentation algorithms with deep learning models are usually very high and pose great computational burden for a lot of applications.

Compared to 3D data segmentation, 2D image segmentation has been much better addressed. There are many well-established algorithms for 2D image segmentation. The amount of well annotated 2D image data is magnitude of that of the 3D data, which leads to significantly better deep leaning based 2D image segmentation algorithms. For example, the SAM model released recently by Meta achieves excellent performance in 2D image segmentation tasks[11]. In addition, though there are several different color schemes for 2D images, the transferring between different image file format is a trivial task and most 2D image segmentation methods are thus general applicable and not limited to a specific image format or data model.

For 3D reconstruction studies, existing methods usually rely on 2D image sequences. A typical algorithm is multi-view stereo reconstruction [12], [13], which uses the multiple views (image) of a 3D object to estimate the camera poses by matching pixel RGB information and then project RGB information into 3D space based on estimated camera poses. Due to the advances of image capturing device, the RGBD-based reconstruction method[14] also rapidly gain momentum, which estimate the poses of individual fragments using depth information captured from one view or multiple views, calibrate poses using RGB information, fuse the fragments with RGB information and project them onto a fused 3D scene or model.

In this study, we propose a one-stop 3D target reconstruction and part segmentation framework (OSTRA) with open-source implementation. Our consideration is to integrate pixel-by-pixel segmentation labels of the original 2D images into the reconstruction of the 3D objects. The framework takes full advantage of the advances of the 2D RGB image segmentation technique and projects the segmentation labels in 2D space into 3D space when the RGB information is projected. The reconstructed 3D data thus contains the segmentation labels directly. We note that there are methods for 3D point cloud segmentation, such as projection-based segmentation [15]–[17], which project 3D data to multiple 2D snapshots for segmentation. Our method looks similar but is different. The projection-based segmentation method works on an existing 3D scene and comprises dimension reduction, segmentation and 3D reconstruction again. During this process, the choice of viewpoint in the 3D scene, the loss of information during projection and the inaccuracy of the segmentation all lead to the accumulation of errors. In contrast, OSTRA works on the 2D image sequences directly and integrates the segmentation and reconstruction into a one-stop process. It uses original 2D images as input for segmentation and reconstruction, and is thus free from viewpoint-based dimension reduction and ensures the integrity of the data. In addition, OSTRA analyses the internal structure and relationships in 3D data to prevent errors in data understanding.

OSTRA has three advantages. The first is that it takes advantage of the advances in the 2D image segmentation studies, especially the Segment-Anything Model (SAM) which uses the pretrained neural network without additional training for new scenes, for 3D data segmentation. It saves us from the time-consuming and costly 3D data annotation. The second advantage stems from OSTRA's support of all levels of segmentation, such as semantic segmentation (scene level), instance segmentation (object level) and part segmentation (part level) through text or click prompts. It can handle customized requirement for specific tasks. At last, OSTRA shows promising results for existing 3D datasets especially for structurally complex objects. The segmentation labels by OSTRA are even closer to the ground truth than manually annotated labels for occluded parts in certain experiments.

**METHOD**

OSTRA seamlessly integrates 3D reconstruction technology, 2D object segmentation techniques and object tracking techniques to address the challenging problem of 3D target reconstruction and multilevel segmentation in complex scenes. OSTRA separates three steps (Fig. 1): firstly, it identifies objects with the segment-anything model in the 2D image; secondly, it tracks multiple instances with segmentation labels in the image sequence; at last, it reconstructs labelled 3D objects or multiple parts with Multi-View Stereo (MVS) or RGBD-based 3D reconstruction methods. It supports user interaction through text or click prompts, shown in Fig. 2, to satisfy the customized requirement of the resolution of segmentation.

**Identifying Objects with the Segment-Anything Model**

OSTRA uses the Segment-Anything Model (SAM)[11] to identify and extract the object regions of interest in the input image. SAM takes an image *I* and a prompt *p* that specifies a point or bounding box annotation as input, and outputs accurate segmentation masks *M* as

$$M = SAM(I, p), \quad p = \{p_{text}, p_{click}\}$$

Where the $p_{text}$ is obtained by Grounding-DINO[18] through the input text and the $p_{click}$ is obtained through the user interaction interface (Fig. 2).

**Generating continuous masks for the image sequence with the video object segmentation technique**

We use the video object segmentation technique[19], [20] to generate continuous masks for the image sequence. Given a video, the target of video object segmentation is to predict the mask in the current frame based on the image and the mask in the previous frame, as well as a pre-trained model, as formulated in the following equation:

$$M_t = f(I_t, M_{t-1}, P)$$

where $M_t$ is the target mask in the *t*-th frame of the video (or *t*-th image in the sequence of images), $I_t$ is the image of the *t*-th frame of the video, $M_{t-1}$ is the target mask in the (*t*−1)-th frame of the video, *P* are the model parameters, and *f* is a pre-trained nonlinear function that represents the video object tracking algorithm.

**Segmentation Mask Projection for MVS-based 3D reconstruction**

We develop a novel segmentation mask project algorithm for MVS-based construction. In MVS-based construction, SfM[12] is used to estimate the camera pose for each view (image), and it comprises five steps: feature extraction, feature matching, camera pose estimation, triangulation and global optimization. The SfM process outputs the extrinsic parameters *R*,*t*, the intrinsic parameter *K*, and the sparse point cloud $P_{\text{sparse}}$, described as follows:

$$R, t, K, P_{\text{sparse}} = SfM(I_{RGB})$$

where $I_{RGB}$ are the RGB images. Then, the PatchMatch Stereo (PMS) algorithm[25] is used to estimate the depth map $D$ and the normal map $N$ for each view, formulated as:

$$D, N = PMS(I_{RGB}, R, t, K, P_{\text{sparse}})$$

Depth map pixels are then represented in 3D space with following equation:

$$\begin{bmatrix} X \\ Y \\ Z \\ W \end{bmatrix} = dK^{-1} \begin{bmatrix} u \\ v \\ 1 \end{bmatrix}, \begin{bmatrix} x \\ y \\ z \end{bmatrix} = \frac{1}{W} \begin{bmatrix} X \\ Y \\ Z \end{bmatrix}$$

where ($u$,$v$) are the 2D coordinates of the pixel in the view and the depth map, ($X$,$Y$,$Z$,$W$) are the homogeneous coordinates of the pixel ($u$,$v$), ($x$,$y$,$z$) are the 3D coordinates of the pixel ($u$,$v$), and $d$ is the depth value corresponding to the pixel.

In MVS-based construction, $D$, $N$ and other parameters are estimated from $I_{RGB}$ to produce the final point cloud. The depth map $D$ describes the position relative to the camera and the normal map $N$ is used for the subsequent optimization. We extend the above construction method for the segmentation label $I_{mask}$. Since $I_{mask}$ and $I_{RGB}$ use the same 2D coordinate, a mask-based point cloud can be produced with $I_{mask}$ using parameters estimated from $I_{RGB}$. The 3D reconstruction process of OSTRA is formulated as follows:

$$P_{RGB} = fusion(D, N, R, t, K, I_{RGB})$$

$$P_{mask} = fusion(D, N, R, t, K, I_{mask})$$

$P_{mask}$ and $P_{RGB}$ contain the same number of data points and the corresponding data point has the same coordinate in the 3D space. The final 3D object with segmentation labels can be obtained by integrating information of $P_{mask}$ and $P_{RGB}$ and filtering out the inconsistent points caused by noise as follows:

$$P = filter(P_{RGB}, P_{mask})$$

The output $P$ is a point cloud with one or more segmentation targets.

**Segmentation Mask Projection for RGBD-based 3D reconstruction**

We develop a new segmentation mask project algorithm for RGBD-based 3D reconstruction. The RGB information is mapped onto the scalable TSDF voxel grid together with the mask information to create a rich 3D geometry with segmentation information. The resulting 3D model can be represented as a point cloud, mesh, or voxel.

The process of fragment generation creates fragments from the RGBD image sequence. Assume *G* is a set of poses, and *P* is a set of point clouds for each fragment. *C* is the collection of color images from the RGBD sequence, and *D* is a collection of depth images. *K* is the camera intrinsic parameter. The fragment generation process is defined as follows:

$$G, P = MakeFragments(C, D, K)$$

Next, the point cloud for each fragment is pre-processed, including down-sampling and feature extraction based on their poses. For each pair of adjacent fragments, the transformation matrix between them is calculated using either Fast Global Registration[22] based on feature matching or RANSAC[23] based on global registration. Subsequently, the transformation matrices between each pair of adjacent fragments are refined using multiscale ICP[24] to improve the registration accuracy. By denoting the resulting set of poses after registration as *G′*, the process can be expressed as:

$$G' = RegisterFragments(G, P)$$

The integration process is then applied to create a scalable TSDF volume to fuse the fragments based on their poses and the information of the image sequence. We create a new data structure MASKD, which consists of a mask image and a depth image. In the same way as the RGBD sequence is used for reconstruction, the MASKD sequence can also be used to construct a TSDF volume:

$$V_{mask} = Integration(M, D, K, G', P)$$

$$V_{RGB} = Integration(C, D, K, G', P)$$

here *M* is the collection of mask images from the MASKD sequence. TSDF volumes $V_{mask}$ and $V_{RGB}$ are constructed using the MASKD and RGBD sequences respectively, and are structurally identical. TSDF volumes are further integrated to produce specific data models in voxel, mesh or point cloud.

**Implementation of the method**

Grounding-DINO[18] (pre-trained model groundingdino_swint_ogc) is used to support the input of the text prompt for image segmentation . We use the Segment-Anything[11] Model (SAM, pre-trained model sam_vit_b_01ec64) to segment the first frame of the video. DeAOT[19], [25] (pre-trained model R50_DeAOTL_PRE_YTB_DAV), Xmem[20], [26] (pre-trained model XMem-s012) are used to track video targets and generate masks. We use SuperPoint[27], SuperGlue[28] for feature extraction and matching, pyColmap[12], [13], hloc[29] toolkit for SfM implementation. Colmap [12], [13] is used to implement the PatchMatch Stereo and Fusion algorithms. Open3D reconstruction system[14] is used to implement the RGBD fragment generation, registration and integration. OSTRA is developed in Python.

**RESULTS**

We applied OSTRA to several publicly available datasets. We also produced several practical datasets to investigate the performance of our method. In the MVS-based segmentation, five public MVS datasets (DTU Robot Image Data Sets[30], Middlebury MVS Data Sets[31], Tanks and Temples Data Sets[32], ETH3D Data Sets[33] and Redwood Data Sets[34]) were chosen. In addition, three datasets with the image sequences of Lego objects and a database with a rice tiller image sequence were produced and tested. In the RGBD-based experiments, we conducted segmentation on three public RGBD datasets (Redwood Data Sets[34] , SceneNN Data Sets[35] and Stanford 3D Indoor Scene Dataset[36]). These datasets include indoor scenes, outdoor scenes and close-up scenes, covering different lighting conditions, camera angles and image resolutions.

OSTRA provides a visual interface for monitoring levels of segmentation. It supports the text prompt or the mouse click. We used Meta's Segment Anything Model (SAM) for the interactive segmentation. The first frame of the video is segmented using interactive segmentation. OSTRA then loads the configured Video Object Segmentation (VOS) model, either "DeAOT" (Decoupling Features in Hierarchical Propagation for Video Object Segmentation) or "XMem" (Long-Term Video Object Segmentation with an Atkinson-Shiffrin Memory Model) for object tracking. During video tracking, there are occasionally tracking errors due to lighting, occlusion or blur, and interactive segmentation is activated for correction to ensure the accuracy of the segmentation. After the tracking process is completed, OSTRA outputs a set of mask images together with the segmentation labels for 3D reconstruction. In the MVS-based case, OSTRA loads the reconstruction frame configuration and takes a reconstruction frame every $x$ tracking frames (the smaller $x$, the denser the reconstructed point cloud and the higher time cost and GPU usage), while in the RGBD-based case, OSTRA uses all frames for reconstruction.

Unfortunately, as OSTRA integrates the segmentation with 3D reconstruction from the image sequence, rather than segment the point cloud data directly as the existing methods do, it is very challenging if not impossible to use ground truth from the public dataset for quantitative comparison. We have verified the feasibility of our method through a series of qualitative visualization results. An PC with an AMD Ryzen 7 5800 8-core processor CPU and a NVIDIA RTX 3060 GPU is used for testing. The experimental results are shown in Tables 1, 2 and Figures 4, 5. Interestingly, DeAOT achieves significantly higher segmentation performance and efficiency than XMem for the Lego dataset. But for other datasets, XMem achieves comparable or higher performance and efficiency than DeAOT. This justifies OSTRA's support to both the XMem-based and the DeAOT-based tracking algorithms.

Table 1: The statistics of OSTRA reconstruction and segmentation for MVS datasets, with the visual performance shown in Fig. 4.

| Number | Image set | Resolution | Tracking frames | Tracking time (s) | | Reconstruction frames | Reconstruction time (s) | Segmentation Type |
|---|---|---|---|---|---|---|---|---|
| | | | | DeAOT | XMem | | | |
| 1 | Temple[31] | 640×480 | 312 | 49.83 | 32.08 | 23 | 180.07 | part |
| 2 | Tank[32] | 1024×540 | 562 | 107.71 | 70.8 | 53 | 959.58 | instance |
| 3 | Horse[32] | 960×540 | 601 | 115.46 | 90.69 | 61 | 1104.12 | instance |
| 4 | Livingroom[34] | 640×480 | 2870 | 420.71 | 304.847 | 139 | 2221.07 | semantic |
| 5 | Courtyard[33] | 1512×1008 | 15 | 13.24 | 13.44 | 12 | 325.88 | semantic |
| 6 | Scan1[31] | 800×600 | 49 | 12.33 | 6.75 | 49 | 1008.12 | instance |
| 7 | Scan6[31] | 800×600 | 49 | 10.29 | 6.77 | 49 | 1067.33 | instance |
| 8 | Lego-car | 1280×720 | 624 | 196.4 | 486.26 | 24 | 763.07 | part |
| 9 | Lego-person | 720×1280 | 1377 | 408.83 | 970.41 | 137 | 3954.24 | part |
| 10 | Lego-scence | 1280×720 | 908 | 238.56 | 407.88 | 38 | 1139.86 | semantic |
| 11 | Rice tiller | 960×540 | 195 | 37.25 | 19.70 | 195 | 2974.06 | part |

Table 2: The statistics of OSTRA reconstruction and segmentation for RGBD datasets, with the visual performance shown in in Fig. 5.

| Number | Image set | Resolution | Tracking frames | Tracking time (s) | | Reconstruction frames | Reconstruction time (s) | Segmentation Type |
|---|---|---|---|---|---|---|---|---|
| | | | | DeAOT | XMem | | | |
| 1 | Livingroom [34] | 640×480 | 2870 | 457.84 | 449.23 | 2870 | 926.93 | instance |
| 2 | Lounge[36] | 640×480 | 3000 | 485.29 | 233.77 | 3000 | 227.59 | instance |
| 3 | SceneNN-337[35] | 640×480 | 1542 | 254.48 | 101.99 | 1542 | 519.31 | semantic |
| 4 | SceneNN-016[34] | 640×480 | 1364 | 227.58 | 139.37 | 1364 | 411.49 | instance |
| 5 | SceneNN-021[34] | 640×480 | 3058 | 487.67 | 448.39 | 3058 | 1024.66 | instance |
| 6 | SceneNN-016[34] | 640×480 | 1321 | 239.23 | 186.09 | 1321 | 411.09 | part |

**Semantic segmentation (scene level)**

We take the "lego-scence" image set as an example for illustration (No. 10 in Fig. 4). In this scene, there are a total of 6 wheels, 4 installed on Lego cars and 2 stacked in the scene. We semantically segment all wheels in the scene by keying in the word "wheel" in the prompt box (mouse click is also available). The video tracking model XMem is used. Next, we perform video tracking tasks to generate masks corresponding to video frames. Then, we run OSTRA to generate RGB point clouds and mask point clouds with segmentation labels. The same procedure has been used for all semantic segmenatation experiments (No. 4, 5 in Fig. 4 and No. 3 in Fig. 5).

**Instance segmentation (object level)**

We take the "horse" image set of the Tank and Temple public dataset as an example for illustration (No. 3 in Fig. 4). In this outdoor scene, there are many objects and we try to perform instance segmentation on horses in the scene. We first key in the word "horse" in the prompt box (mouse click is also available), then select the video tracking model XMem. Next, we perform video tracking to generate masks corresponding to video frames. Then, we run OSTRA to generate RGB point clouds and mask point clouds with segmentation labels. The same procedure has been used for all instance-segmentation experiments (No. 2, 6, 7 in Fig. 4 and No. 1, 2, 4, 5 in Fig. 5).

**Part segmentation (part level)**

We take the "lego-car" image set as an example (No. 8 in Fig. 4). In this scene there is only one Lego-assembled car with many parts of different shapes and sizes. We try to part segment all its parts. First, we click on all Lego parts that need to be segmented and select DeAOT as the video tracking model. Next, we perform video tracking to generate masks corresponding to video frames. Then, we run OSTRA to generate RGB point clouds and mask point clouds with segmentation labels. The same procedure has been used for all part-segmentation experiments (No. 1, 8, 9, 11 in Fig. 4 and No. 6 in Fig. 5).

**High Performances of OSTRA**

OSTRA achieves high performance for semantic segmentation, instance segmentation and part segmentation on above 3D datasets. Interestingly, OSTRA even achieved very precise segmentation for scenes which are typically regarded as challenging. For example, when multiple parts to be segmented are tightly intertwined, as in the case of Fig. 3 where multiple parts of the Lego car are so tightly stacked, it is very challenging to perform part segmentation even for the naked eye, but OSTRA's results are generally good as shown in Fig. 3.

Another example is the segmentation of the rice tiller dataset, where we attempted to segment stems, leaves and panicles. We compared the results of OSTRA and these of the deep learning point cloud segmentation framework pointnet++ , using the manual segmentation results as a benchmark. The pointnet++ model uses a 9-channel input sources (coordinates, color information and normal vector information) and is trained with 20 tiller samples. The results of pointnet++ and OSTRA are shown in Table 3 and Fig. 6.

Table 3: The statistics of segmentation results for the rice tiller databset, with the manual segmentation as a benchmark.

|  | OSTRA | | | Pointnet++ | | |
| --- | --- | --- | --- | --- | --- | --- |
|  | Panicle | leaf | stem | panicle | leaf | stem |
| IoU | 0.9998 | 0.9505 | 0.9904 | 0.9740 | 0.1567 | 0.1791 |
| Acc | 0.9974 | | | 0.7594 | | |

The difficulty in segmenting the leaf and stem of rice tiller is due to similar characteristics. In the experimental results of pointnet++, the IoU of the panicle was 0.97, while the IoU of both leaf and stem were less than 0.18. In the experimental results of OSTRA, the IoU of both panicle and stem were above 0.99 and the IoU of the leaf was 0.95. We further investigated the reasons for the slightly lower IoU of the leaf than the panicle. Interestingly, the main discrepancy between OSTRA segmentation and manual segmentation stems from the rice sheath region, which is a special type of leaf and wraps around the stem. In manual segmentation, the point cloud belonging to the sheath cannot be separated from the stem, but in the mask image from OSTRA the sheath is clearly segmented. As shown in Fig. 6, the sheath in OSTRA(part) is labelled as leaf and the sheath in manual(part) is labelled as stem, indicating that the segmentation result of OSTRA are even closer to the ground truth in such complex cases.

**Reconstruction noise suppression by OSTRA**

OSTRA can reduce noise effectively. Most noise of 3D reconstruction comes from incorrect projection of the background and some comes from incorrect estimation of the edge color of objects. Taking the "lego-person" image set as an example, incorrect projection of background causes point cloud noise in lego-person legs, and incorrect estimation of edge color causes contour noise on lego-person edges. During the reconstruction and multilevel segmentation of OSTRA, noise points from the point cloud do not obtain classification labels, making noise filtering easy to implement and with high performance (Fig. 7).

**CONCLUSIONS**

In this study, we propose OSTRA, a one stop 3D target reconstruction and multilevel segmentation framework. It provides a solution for quickly reconstructing 3D targets embedded with rich multi-scale segmentation information in complex scenes. One of the key advantages of OSTRA is its ability to leverage advances in 2D RGB image segmentation techniques. By extending object tracking and 3D reconstruction algorithms to support continuous segmentation labels, OSTRA ensures robust 3D target reconstruction and segmentation. When integrating with SAM 2D segmentation method, our framework essentially eliminates the need for extensive training on 3D scenes, reduces significantly the time, cost and workload of data annotating.

OSTRA provides flexibility and adaptability to specific segmentation tasks. It supports semantic segmentation, instance segmentation and part segmentation at different levels through text or click prompts. Moreover, it can be applied to almost all 3D object data models such as point cloud, mesh and voxel. This allows users to tailor the segmentation process to their specific needs, making OSTRA highly versatile and applicable to a wide range of applications.

The effectiveness of OSTRA is demonstrated by its ability to generate 3D segmentation that are closer to the ground truth than this from the naked eye. This is particularly valuable in cases where manual annotation is difficult due to complex structures or occlusions. In addition, OSTRA provides an intuitive visual interface for interactive segmentation and calibration. This user-friendly interface ensures ease of use and enables researchers to effectively calibrate OSTRA for various applications.

Furthermore, our method is highly modular. In detail, the extended 2D tracking algorithm should be applicable for the tracking of labeled multiple objects or an object with separately labeled multiple parts regardless of the specific algorithm. The extended 3D reconstruction module can be widely applied for generating 3D objects with labeled multiple parts. Although we have focused on one-stop 3D reconstruction and segmentation in this paper, the modules in OSTRA should also work equally well in other applications. For example, it is possible to incorporate OSTRA modules into 3D projection-based segmentation algorithms to improve the performance and efficiency.

Finally, there is still room to improve OSTRA. During our experiments, we found that SAM did not work well for very thin objects, for example rice stems, and the problem was resolved only after we added in several images taking in a much closer range. Similarly, VOS models failed in some cases, which is might due to the fact that they are designed to track dynamic scenes, but are used to track static scenes in our framework. The problem could be resolved in the future by adding the relative positional relationships of the different segmented objects in the static scene to constrain the tracking strategy of the VOS.

**ACKNOWLEDGEMENTS:**

We thank M. Tsiantis for his valuable comments. This work was supported in part by a grant from the National Key R&D Program of China (Grant No.2022ZD0115703), the National Natural Science Foundation of China (Grant No. 32170647), and the grants from the National Science Foundation of Jiangsu Province in China (Grant No. JSSCRC2021508, BE2022383 and BK20212010), and Jiangsu Collaborative Innovation Center for Modern Crop Production.

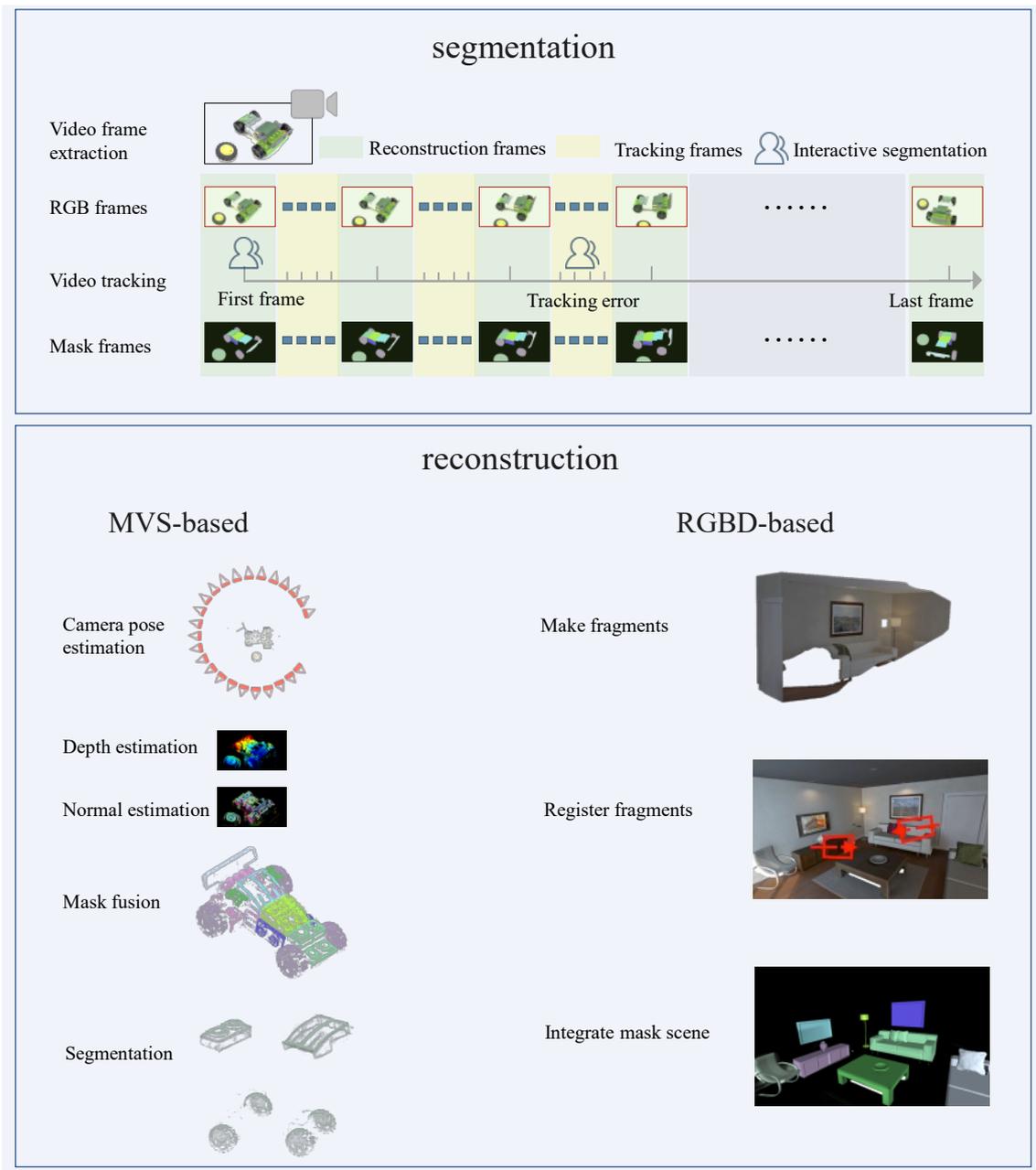

Fig.1: The workflow of OSTRA. The first frame of the video is segmented with the extra support of use interaction and the segmented mask is passed to the VOS model. The VOS model continuously generates masks for subsequent frames. If a tracking error occurs during this process, interactive segmentation can be activated for correction to ensure the integrity of the masks for frames under processing. In the MVS-based case, OSTRA takes a frame for reconstruction among $x$ tracking frames (the smaller $x$, the denser the reconstructed point cloud, and the higher the time cost and GPU usage), and performs camera pose estimation, depth estimation and direction vector estimation according to the reconstruction frame. In the case of RGBD-based reconstruction, OSTRA first converts RGBD images into point cloud fragments, then aligns these fragments in the same 3D space, and finally projects the RGB information and segmentation labels onto the 3D model.

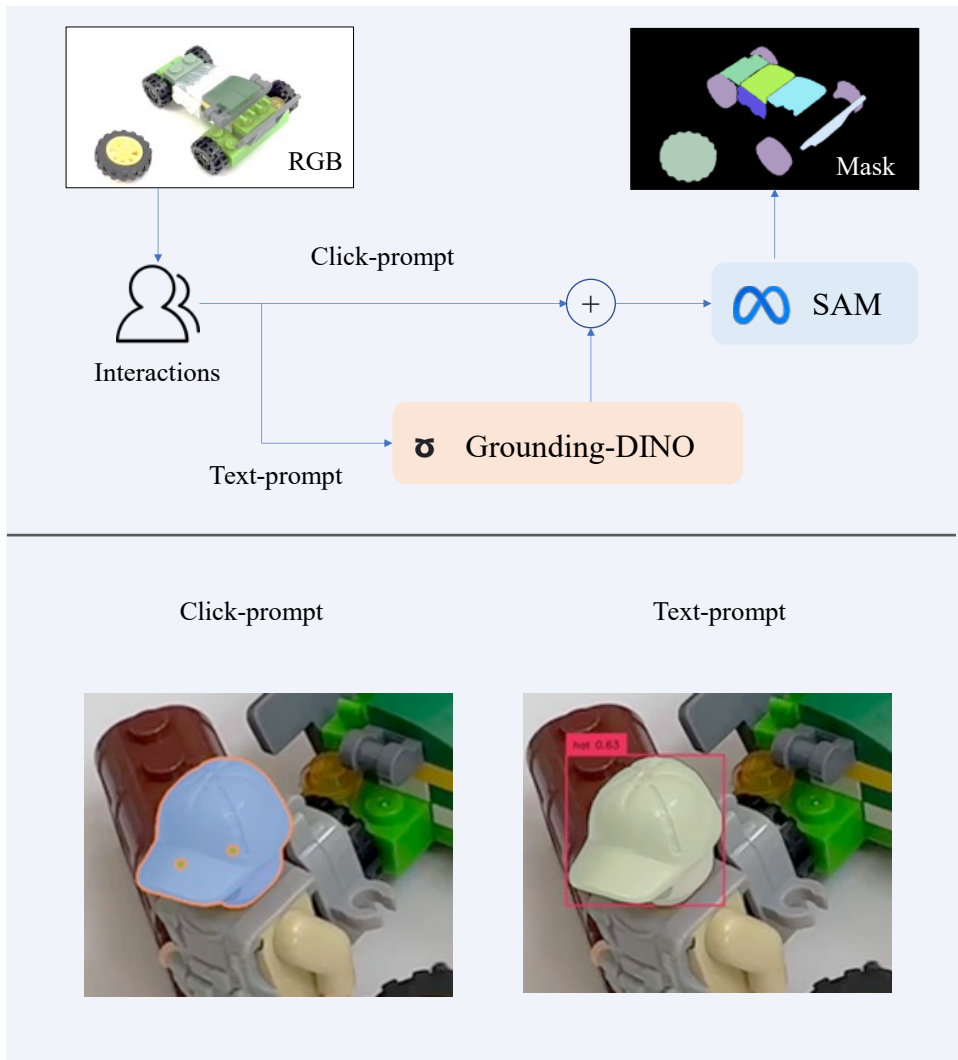

Fig.2: The support of the interactive segmentation. The user selects the object to be segmented in the image frame. The object can be selected by clicking with the mouse or keying in texts in the prompt box for semantic recognition and object locking. Then SAM algorithm is used to segment the object area and generate the segmentation mask.

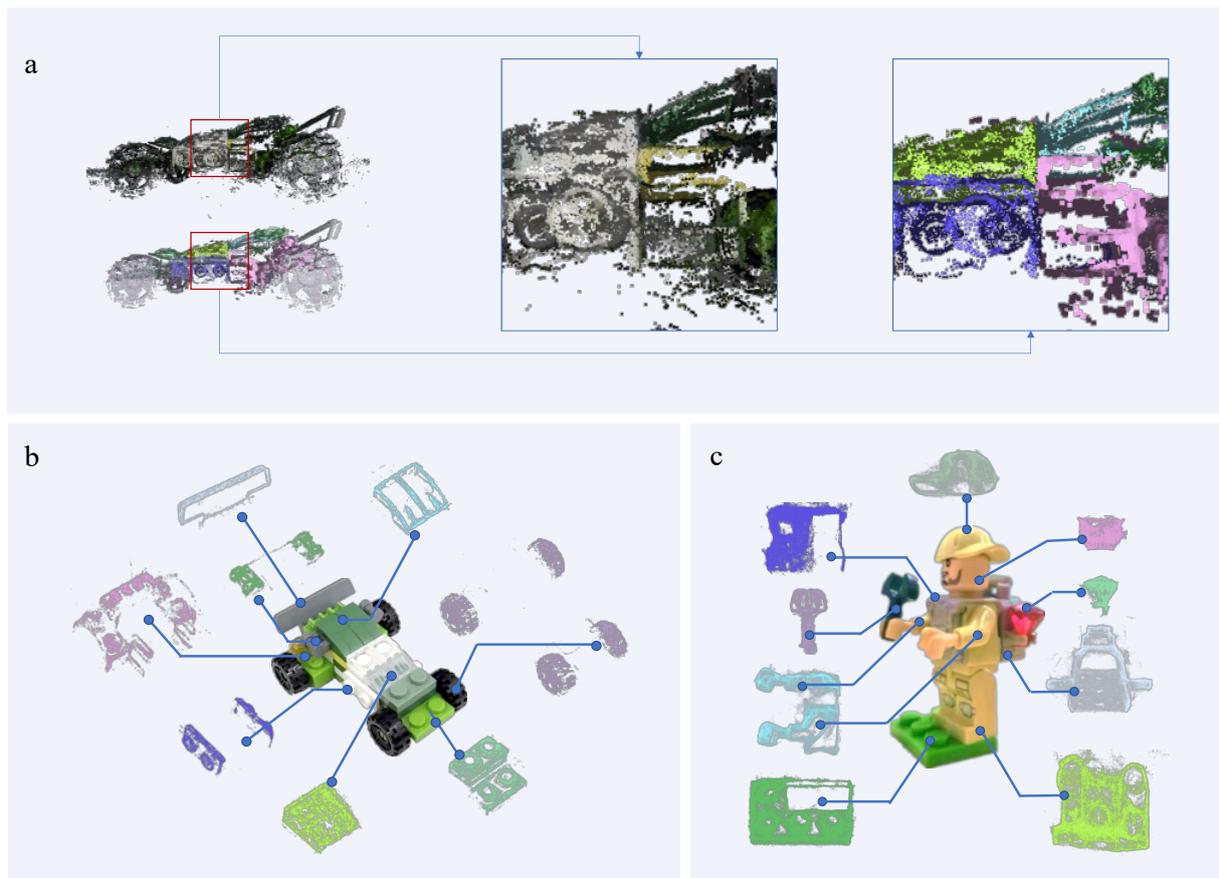

Fig.3: Segmentation of complex objects with subtle structures using OSTRA. (a) OSTRA segmentation labels (the lower panel) are closer to the ground truth than manually annotated labels (the upper panel) where parts are densely packed. (b) Part segmentation of a lego car. (c) Part segmentation of a lego person.

| Number | RGB point cloud | Mask point cloud | Number | RGB point cloud | Mask point cloud |
|---|---|---|---|---|---|
| 1 | 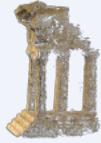 | 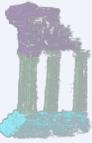 | 2 | 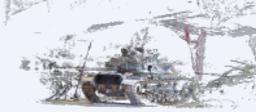 | 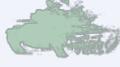 |
| 3 | 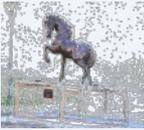 | 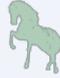 | 4 | 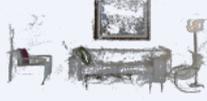 | 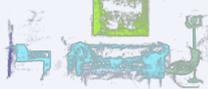 |
| 5 | 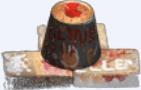 | 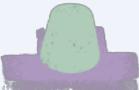 | 6 | 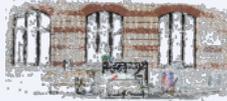 | 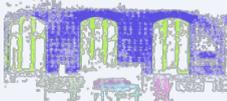 |
| 7 | 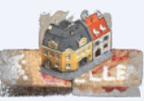 | 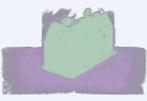 | 8 | 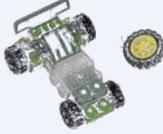 | 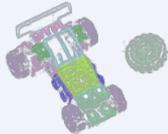 |
| 9 | 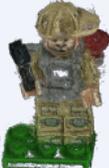 | 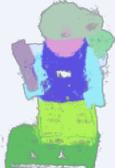 | 10 | 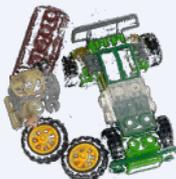 | 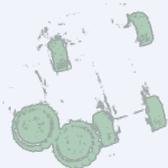 |
| 11 | 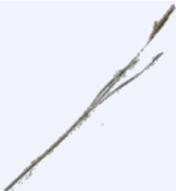 | 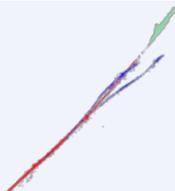 | | | |

Fig.4: The MVS-based OSTRA segmentation. No. 4, 5, 10 are for semantic segmentation, No 2, 3, 6, 7 for instance segmentation, and No. 1, 8, 9,11 for part segmentation. The RGB point cloud shows the reconstruction using RGB information of video frames. The mask point cloud shows the reconstruction with extra information of masks of video frames.

| Number | RGB mesh | Mask mesh | Mask point cloud | Mask voxel |
|---|---|---|---|---|
| 1 | 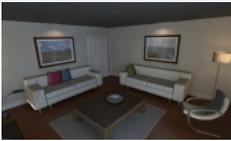 | 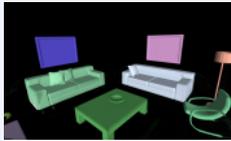 | 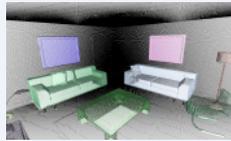 | 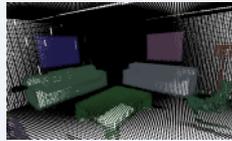 |
| 2 | 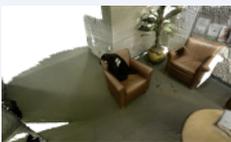 | 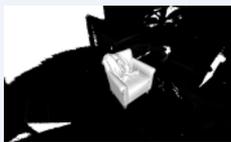 | 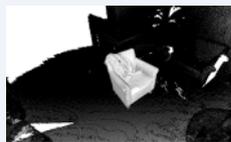 | 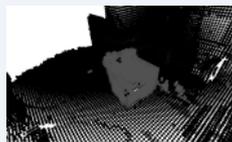 |
| 3 | 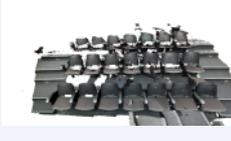 | 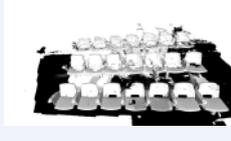 | 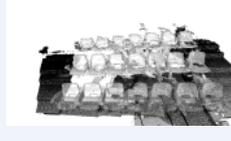 | 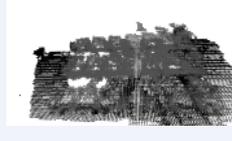 |
| 4 | 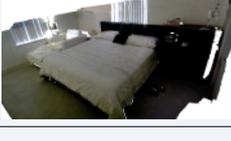 | 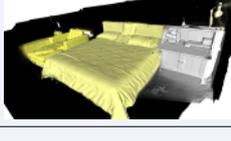 | 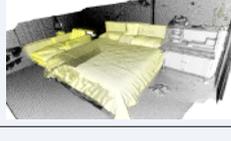 | 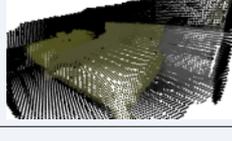 |
| 5 | 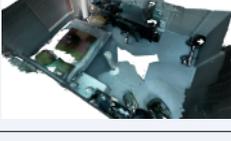 | 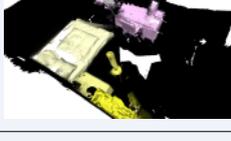 | 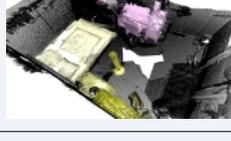 | 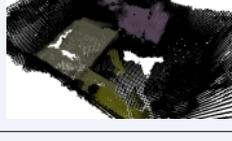 |
| 6 | 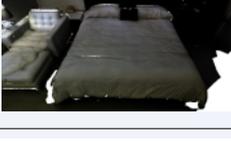 | 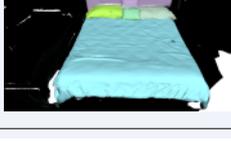 | 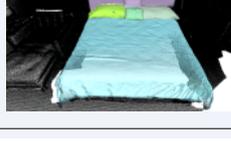 | 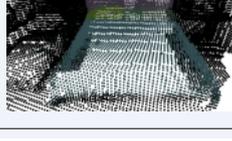 |

Fig.5: The RGBD-based OSTRA segmentation. No. 3 is for semantic segmentation, No. 1, 2, 4, 5 for instance segmentation, No. 6 for part segmentation. The RGB mesh shows the 3D reconstruction of RGBD images. The mask mesh, mask point cloud and mask voxel show the reconstruction with the information of masks of 2D images.

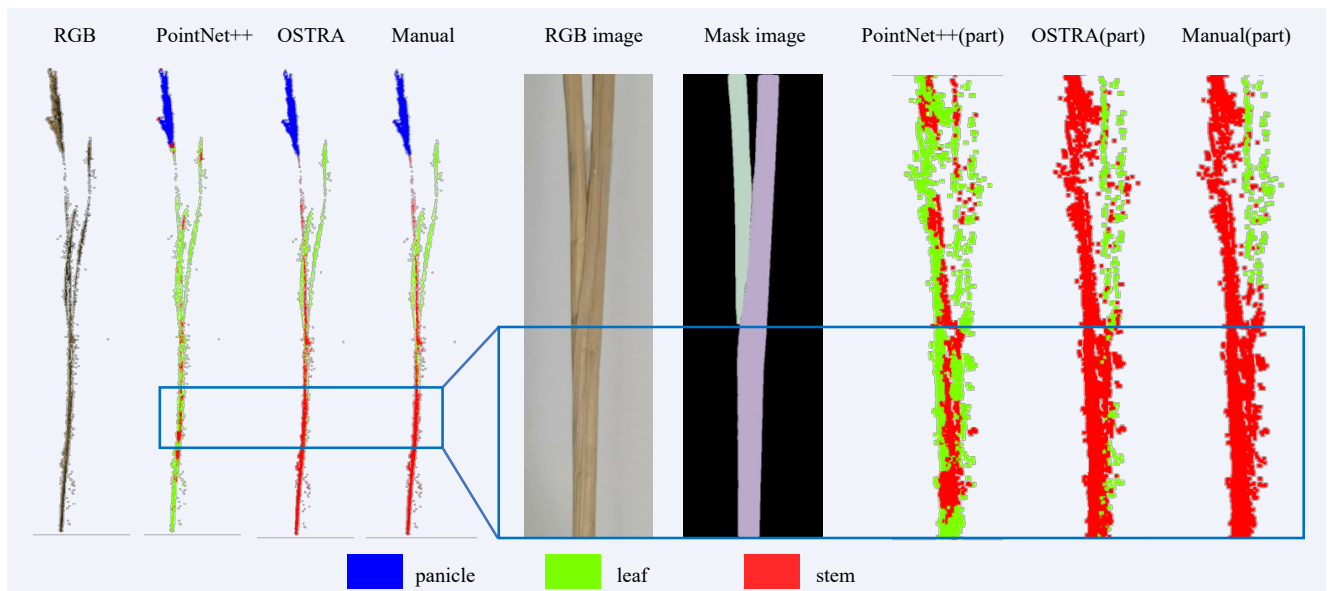

Fig.6: The segmentation of the rice tiller . For pointnet++, the IoU of the panicle was 0.97, while the IoU of both leaf and stem were 0.17. For OSTRA, the IoU of both panicle and stem were above 0.99 and the IoU of the leaf was close to 0.95. In the part segmentation results of OSTRA and manual annotation, the sheath in OSTRA(part) is labelled as the leaf and the sheath in manual(part) is labelled as the stem.

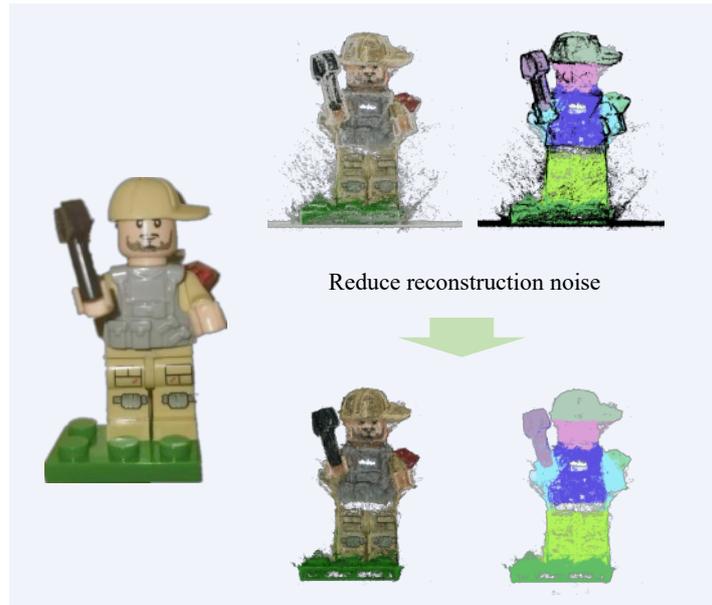

Fig.7: Noise suppression during reconstruction of OSTRA. In the reconstructed point cloud, the points of noise have no segmentation labels and are filtered out by OSTRA.